\documentclass[sigconf,nonacm]{acmart}

\usepackage{tikz}
\usetikzlibrary{arrows.meta,positioning}
\usepackage{booktabs}
\usepackage{amsmath}
\usepackage{adjustbox}
\microtypesetup{expansion=false}
\AtBeginDocument{\Urlmuskip=0mu plus 2mu\relax}

\copyrightyear{2026}
\acmYear{2026}
\setcopyright{cc}
\setcctype{by}
\acmConference[GECCO Companion '26]{Genetic and Evolutionary Computation Conference}{July 13--17, 2026}{San Jose, Costa Rica}
\acmBooktitle{Genetic and Evolutionary Computation Conference (GECCO Companion '26), July 13--17, 2026, San Jose, Costa Rica}
\acmDOI{10.1145/3795101.3814645}
\acmISBN{979-8-4007-2488-6/2026/07}

\makeatletter
\renewcommand{\@copyrightpermission}{%
  \textbf{CC-BY}\\[2pt]
  \href{https://creativecommons.org/licenses/by/4.0/legalcode}{%
    \includegraphics[height=5ex]{doclicense-CC-by-88x31}}\\[2pt]
  \href{https://creativecommons.org/licenses/by/4.0/legalcode}{%
    This work is licensed under a Creative Commons Attribution International 4.0 License.}\\[4pt]
  \textit{GECCO Companion '26 July 13--17, 2026, San Jose, Costa Rica}\\
  \textcopyright\ 2026 Copyright is held by the owner/author(s).\\
  ACM ISBN 979-8-4007-2488-6/2026/07\\
  \href{https://doi.org/10.1145/3795101.3814645}{https://doi.org/10.1145/3795101.3814645}%
}
\makeatother

\begin{document}

\title{Genesis: An Empirical Platform for Studying Open-Ended
       Evolution Without Fitness Functions}

\author{Anushka Sharma}
\email{anushka.care@gmail.com}
\orcid{0009-0002-8158-3748}
\affiliation{%
  \institution{Banasthali Vidyapith}
  \city{Vanasthali}
  \state{Rajasthan}
  \country{India}
}

\begin{abstract}
Biological evolution sustains complex dynamics without any fitness
function, yet virtually all evolutionary algorithms depend on one.
\textbf{Genesis} is an open-source platform designed to test, empirically,
what an artificial system needs to sustain evolutionary dynamics after
complete fitness removal. Evolution in Genesis is governed by physical
constraints, relational dominance, and adaptive regulation---no scalar
fitness, no designer-specified objectives. Across experiments totalling
over one million evolutionary generations, Genesis has:
(\textit{i})~shown that constraint-driven selection can sustain
evolutionary activity after complete fitness removal (7/12 runs; Wilson
95\% CI [30.2\%, 82.5\%]; $p{<}0.01$, Cohen's $d{=}1.47$ vs.\
baselines); (\textit{ii})~produced a sham-controlled negative result
demonstrating that niche construction alone does not break the complexity
plateau; and (\textit{iii})~provided preliminary evidence that
speciation-protected niche construction initiates structural
diversification that unprotected secretion cannot. These findings
establish empirical boundaries for fitness-free evolution and open a new
direction: \emph{meta-evolution of physics}, in which the laws governing
an evolutionary system are themselves evolved.
\end{abstract}



\maketitle

\section{Motivation}

Evolutionary computation (EC) has produced powerful optimisers, but
virtually every EC algorithm shares one assumption: a fitness function
scores solutions and guides selection. Biological evolution has no
such function, yet generates the sustained novelty that defines
open-ended evolution (OEE)~\cite{packard2019}. No prior work has
empirically characterised what happens when fitness is
\emph{completely removed} or which architectural components determine
whether dynamics survive. Genesis is a \emph{scientific instrument}
built to falsify these hypotheses~\cite{lehman2011,mouret2015,wang2019}.

\section{The Genesis Architecture}

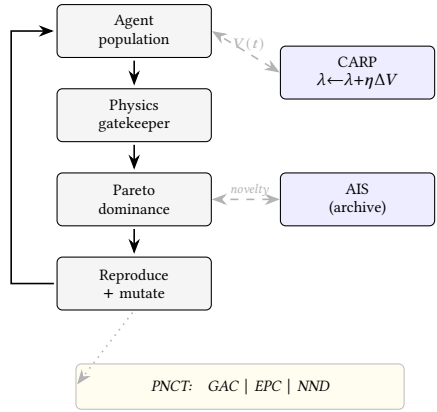
\begin{figure}[t]
\centering
\begin{tikzpicture}[
  node distance=4mm and 9mm,
  box/.style={draw, rounded corners=2pt, fill=gray!8,
              font=\scriptsize, text width=19mm, align=center,
              inner sep=2pt, minimum height=7mm},
  sbox/.style={draw, rounded corners=2pt, fill=blue!7,
               font=\scriptsize, text width=19mm, align=center,
               inner sep=2pt, minimum height=7mm},
  mon/.style={draw=gray!50, rounded corners=2pt, fill=yellow!7,
              font=\scriptsize\itshape, text width=42mm, align=center,
              inner sep=2pt, minimum height=6mm},
  arr/.style={-Stealth, semithick, shorten >=1pt, shorten <=1pt},
  darr/.style={Stealth-Stealth, semithick, dashed, gray!60}
]
\node[box] (pop)  {Agent\\population};
\node[box, below=4mm of pop]  (gk)  {Physics\\gatekeeper};
\node[box, below=4mm of gk]   (par) {Pareto\\dominance};
\node[box, below=4mm of par]  (rep) {Reproduce\\$+$ mutate};
\node[sbox, right=9mm of pop, yshift=-5.5mm] (carp)
      {CARP\\$\lambda{\leftarrow}\lambda{+}\eta\Delta V$};
\node[sbox, right=9mm of par] (ais)  {AIS\\(archive)};
\node[mon, below=7mm of rep, xshift=14mm] (pnct)
      {PNCT:\quad GAC $|$ EPC $|$ NND};
\draw[arr] (pop)--(gk);
\draw[arr] (gk)--(par);
\draw[arr] (par)--(rep);
\draw[arr] (rep.west) -- ++(-6mm,0) |- (pop.west);
\draw[darr] (carp.west) -- (pop.east)
  node[pos=0.45, above, font=\tiny\itshape, inner sep=1pt]{$V(t)$};
\draw[darr] (ais.west) -- (par.east)
  node[pos=0.45, above, font=\tiny\itshape, inner sep=1pt]{novelty};
\draw[arr, dotted, gray!60] (rep.south) -- (pnct.west);
\end{tikzpicture}
\caption{Genesis architecture: main loop (solid), CARP/AIS feedback
(dashed), PNCT monitoring only (dotted---no selection signal).}
\label{fig:arch}
\end{figure}

Genesis replaces fitness-based selection with three interacting
mechanisms (Figure~\ref{fig:arch}), none of which supplies a scalar
reward; all regulate \emph{viability and diversity}, not performance.

\textbf{(1) Constraint-Driven Selection (CDS).}
Agents carry variable-length, codon-based genomes. Survival depends
solely on satisfying immutable physical constraints: a metabolic cost
\[
  M(G)=\alpha|G|^{\beta}+\textstyle\sum_k \gamma_k f_k(G)
\]
penalises genome bloat ($\alpha{=}0.005$, $\beta{=}1.5$). A \emph{Physics
Gatekeeper} enforces hard resource limits; survivors are chosen by
Pareto relational dominance across internal axes (energy efficiency,
resource acquisition, interaction stability)~\cite{deb2001}.

\textbf{(2) Constraint-Adaptive Regulation Principle (CARP).}
CARP dynamically adjusts constraint intensity $\lambda$ to keep
population viability $V(t)$ within a target corridor (70--90\%):
$\lambda(t{+}1) = \lambda(t) + \eta\,(V_{\text{target}} - V(t))$.
CARP regulates \emph{capacity} without encoding any preference over
which behaviours survive~\cite{coello2002}.

\textbf{(3) Artificial Immune System (AIS).}
The AIS maintains a novelty-based archive of representative genotypes,
pruning dominant lineages and reintroducing archived genotypes when
novelty falls below $N_{\min}$.

\textbf{Instrumentation: PNCT Metrics.}
Three falsifiable metrics---the Population-level Non-fitness
Characterisation Toolkit (PNCT)---quantify evolutionary activity
without fitness. \textit{Genetic Activity Coefficient} (GAC): fraction
of genome edits persisting beyond a 500-generation horizon; a run is
declared \textbf{sustained} when median GAC ${\geq}0.1$.
\textit{Expressed Phenotype Complexity} (EPC): LZW-based compression
complexity of expressed phenotypes; positive EPC slope is the stricter
criterion for \textbf{open-ended} growth---a bar no experiment has
yet cleared. \textit{Novelty Network Density} (NND): local density
in phenotype feature space.
Together, GAC tracks whether evolution is \emph{doing anything};
EPC tracks whether it is \emph{going anywhere}; NND tracks whether
the population remains \emph{diverse enough} for either to be meaningful.

\section{Empirical Results: Three Experiments}

\subsection{Experiment 1 - Fitness Removal (V2)}

Twelve independent 10,000-generation runs with progressive elimination
of external fitness ($p_m{=}0.01$, $N{=}512$). \textbf{Result:}
7/12 runs (58.3\%) sustained evolutionary activity after complete
fitness removal (Wilson 95\% CI [30.2\%, 82.5\%]), significantly
outperforming four baselines---Random Search, Fixed Constraints,
Novelty Search, and MAP-Elites ($p{<}0.01$, Cohen's $d{=}1.47$,
Mann-Whitney $U$). Ablation confirms both CARP and AIS are critical:
removing either raises failure rates from 41.7\% to over 90\%;
formal CIs are not yet computed.

\textbf{Critical finding:} EPC plateaued at 140--155 in every
successful run---a structural ceiling confirmed via ablation.
Sustained activity without fitness is achievable, but not automatically
\emph{open-ended}.

\subsection{Experiment 2 - Sham-Controlled Niche Construction (V3)}

We hypothesised that organism-writable environments would break
the EPC ceiling~\cite{odlingsmee2003}. Agents received a
\textit{secretion} action depositing a persistent chemical $S$ into
a $3{\times}3$ neighbourhood (half-life exceeding agent lifespan).
A \textbf{sham control}---secretion executes and cost is deducted,
but $S$ is not updated---isolates niche construction from its
energetic cost. We ran 20 runs (10 real, 10 sham) of 50,000
generations each (1 million total).

\begin{table}[h]
\small\centering
\caption{V3 sham-controlled results (means over 10 runs each).}
\label{tab:v3}
\begin{tabular}{lcc}
\toprule
Metric & Real & Sham \\
\midrule
$S_{\text{mean}}$ (chemical field) & 0.48 & 0.00 \\
LZ76 complexity (action traces)    & 0.068 & 0.068 \\
EPC (phenotype complexity)         & 0.00  & 0.00  \\
\bottomrule
\end{tabular}
\end{table}

Secretion physics functioned correctly, yet EPC remained zero in
both conditions and LZ76 complexity was identical (0.068):
agents' \emph{internal policies} did not respond to the chemical
field. Environment modification without frequency-dependent selection
is insufficient to drive open-ended growth.

\subsection{Experiment 3 - Speciation-Protected Niche Construction
  (V4 Stage~3)}

V3 identified the missing ingredient: novel structures need
protection long enough to accumulate. We therefore added CPPN genomes
and NEAT-style speciation~\cite{stanley2002}. Early validation runs
(two seeds, 500 generations) show real and sham conditions diverging
in genome size and species count where V3 showed none---consistent
with the hypothesis but not yet sufficient to establish it. This
serves as directional scaffolding for the medium-scale replication
currently underway.

\section{Outlook}

The EPC plateau across V2--V3 suggests fixed physics may be
structurally unable to support OEE. Genesis V4 therefore treats
physics itself as evolvable: worlds parameterised by a vector
$\theta$ (diffusion, decay, energy injection) reproduce in
proportion to sustained complexity. A 100-world experiment is in
preparation. Genesis contributes to EC: portable PNCT metrics, a
reusable sham-control protocol, and the hypothesis that substrate
evolvability---not just population dynamics---may be necessary for
open-endedness~\cite{packard2019}.

Code and data: \begin{sloppypar}\url{https://github.com/gearupsmile/genesis-emergence}\end{sloppypar}



\begin{thebibliography}{8}

\bibitem{coello2002}
C.~A.~Coello Coello. 2002.
Theoretical and numerical constraint-handling techniques used with
evolutionary algorithms: a survey.
\textit{Comp.\ Methods Appl.\ Mech.\ Eng.}\ 191, 11--12, 1245--1287.

\bibitem{deb2001}
K.~Deb. 2001.
\textit{Multi-Objective Optimization Using Evolutionary Algorithms}.
Wiley.

\bibitem{lehman2011}
J.~Lehman and K.~O.~Stanley. 2011.
Abandoning objectives: Evolution through novelty alone.
\textit{Evolutionary Computation} 19, 2, 189--223.
\url{https://doi.org/10.1162/EVCO_a_00025}

\bibitem{mouret2015}
J.-B.~Mouret and J.~Clune. 2015.
Illuminating search spaces by mapping elites.
\textit{arXiv}:1504.04909.

\bibitem{odlingsmee2003}
F.~J.~Odling-Smee, K.~N.~Laland, and M.~W.~Feldman. 2003.
\textit{Niche Construction: The Neglected Process in Evolution}.
Princeton Univ.\ Press.

\bibitem{packard2019}
N.~Packard \textit{et~al.} 2019.
An overview of open-ended evolution.
\textit{Artificial Life} 25, 2, 93--103.
\url{https://doi.org/10.1162/artl_a_00291}

\bibitem{stanley2002}
K.~O.~Stanley and R.~Miikkulainen. 2002.
Evolving neural networks through augmenting topologies.
\textit{Evolutionary Computation} 10, 2, 99--127.
\url{https://doi.org/10.1162/106365602320169811}

\bibitem{wang2019}
R.~Wang, J.~Lehman, J.~Clune, and K.~O.~Stanley. 2019.
Paired open-ended trailblazer (POET).
\textit{arXiv}:1901.01753.

\end{thebibliography}
\end{document}